\documentclass{ipi}

\usepackage{times}
\usepackage{epsfig}
\usepackage{graphicx}
\usepackage{amsmath}
\usepackage{amssymb}
\usepackage{enumerate}
\usepackage{setspace,url}
\usepackage{algorithm,algorithmic}
\usepackage{subfigure}
\usepackage{url}
\usepackage{hyperref}
\usepackage[usenames,dvipsnames]{pstricks}

\def\currentvolume{X}
\def\currentissue{X}
\def\currentyear{200X}

\def\ppages{X--XX}

\title[Non rigid geometric distortions correction]{Non rigid geometric distortions correction - Application to atmospheric turbulence stabilization}
\author[Yu~Mao and~J\'er\^ome~Gilles]{}


\begin{document}

\maketitle
\centerline{\scshape Yu Mao}
\medskip
{\footnotesize
 \centerline{Institute for Mathematics and Its Applications, University of Minnesota}
   \centerline{425 Lind Hall 207 Church Street SE, Minneapolis, MN 55455-0134, USA}
} 

\medskip
\centerline{\scshape J\'er\^ome Gilles}
\medskip
{\footnotesize
 \centerline{Department of Mathematics, University of California Los Angeles}
   \centerline{520 Portola Plaza, Los Angeles, CA 90095-1555, USA}
} 
\bigskip

\begin{abstract}
A novel approach is presented to recover an image degraded by atmospheric turbulence. Given a sequence of frames affected by turbulence, we construct a 
variational model to characterize the static image. The optimization problem is solved by Bregman Iteration and the operator splitting method. 
Our algorithm is simple, efficient, and can be easily generalized for different scenarios.
\end{abstract}

\section{Introduction}
\begin{figure}[h]
\begin{center}
\includegraphics[width=0.32\textwidth]{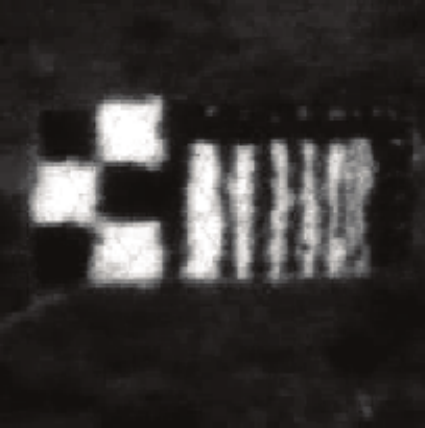}\,
\includegraphics[width=0.32\textwidth]{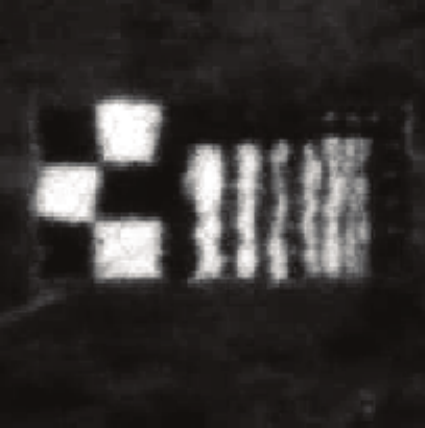}\,
\includegraphics[width=0.32\textwidth]{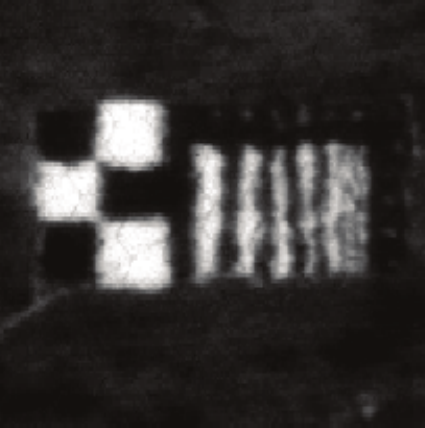}\\
\includegraphics[width=0.32\textwidth]{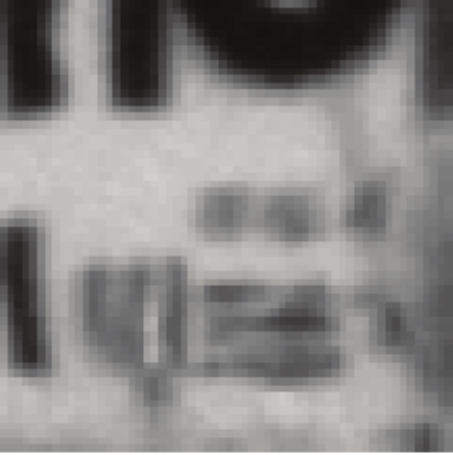}\,
\includegraphics[width=0.32\textwidth]{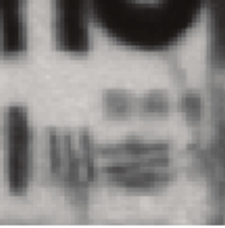}\,
\includegraphics[width=0.32\textwidth]{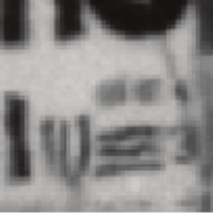}\\
\includegraphics[width=0.32\textwidth]{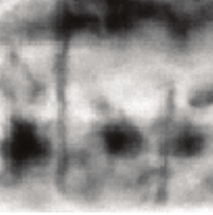}\,
\includegraphics[width=0.32\textwidth]{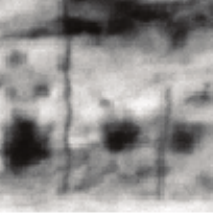}\,
\includegraphics[width=0.32\textwidth]{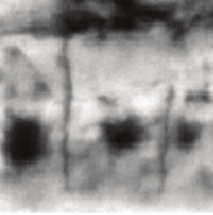}
\end{center}
\caption{Sample images. Each row contains three arbitrary frames from different testing turbulence videos.}\label{videoexample}
\end{figure}

In the last decade, long range imaging systems have been developed to improve target identification. One of the main visual effects is distortion due to 
atmospheric turbulence (known in the literature as ``image dancing''). It may occur in many other scenarios: for example, underwater imaging systems, which are 
subject to scattering effects and video shooting in the summer, suffers from hot air near the ground, and so on. Weak turbulence does not really affect human 
observers, but it can cause problems for an automatic target recognition algorithm because the shape of the object may be very different from those learned 
by the algorithm. Fig.~\ref{videoexample} shows some examples obtained by a camera in real scenarios. For each video we arbitrarily choose three frames to 
display here. 

Previous methods have been developed to deal with the turbulence effect in astronomical images. In \cite{Lemaitre:2007p7023}, local filters 
(Wiener filter, Laplacian regularization and so on) were utilized and local properties were obtained by block partitioning of the image. As a result, 
some block artifacts appear on the restored images.

An interesting work about turbulence modelization for mitigation algorithms was made by Frakes \cite{Frakes:2001p7017,Gepshtein:2004p7030}. The authors 
modeled the turbulence phenomenon by using two operators:
\begin{equation}\label{eq:turbulence}
f_i(x)=D_i(H(u(x)))+\text{noise}
\end{equation}
where $u$ is the static original scene we want to retrieve, $f_i$ is the observed image at time $i$, $H$ is a blurring kernel, and $D_i$ is an operator 
which represents the geometric distortions caused by the turbulence at time $i$. Based on this model, the authors of \cite{Gilles:2008p4901} proposed a 
scheme to evaluate the $H^{-1}$ and $D^{-1}$ operators. The $H^{-1}$ operator is obtained by blind deconvolution, while the correction of the geometrical 
distortions  $D^{-1}$  is computed by an elastic registration algorithm based on diffeomorphic mappings. This approach gives good results but it has two 
main drawbacks. First, note that it is time consuming to perform the calculations due to the two iterative processes involved in the algorithm. Secondly, the 
performance is sensitive to the choice of the parameters.

Another kind of approach for this problem is to utilize the Kalman filter, which is a statistical tool that recovers a static object from a time series of 
observations. In \cite{Tahtali:2008p7021}, the authors successfully use this filter in the turbulence reconstruction problem. However, this method requires 
a strong time dependence of the frames, therefore the frame rate has to be sufficiently high. It treats the warped frames ordered in time as governed by 
fluid dynamics, and thus can be characterized by time-dependent differential equations, which is not a practical assumption in some applications. 

More recently, some efforts were made to propose new mitigation algorithms. In \cite{Li2007}, assuming long exposure video capture, the authors propose to 
use Principal Component Analysis to find the statistically best restored image from a sequence of acquired frames. In \cite{Aubailly2009}, the authors 
use the assumption that for a fixed location in the image, its neighborhood has some high probability to appear with better quality through the time. Then 
the restored region is a fusion of the best ones. Some spatially variant deblurring was proposed in \cite{Hirsch2010} but the algorithm does not specifically 
address the problem of geometrical distortions. Following the modelization of Frakes \cite{Frakes:2001p7017}, in \cite{Zhu2010} the authors propose to invert 
the geometric distortions and the blur. They use some B-Spline registration algorithm embedded in a Bayesian framework with bilateral total variation (TV)
regularization.\\

The goal of this paper is to propose a new approach on this problem, principally on the correction of geometrical distortions. We develop a unified 
framework which uses both an optical flow scheme to estimate the geometrical distortion, and a nonlocal TV based regularization process to recover the 
original observed scene. The paper is organized as follows: in section \ref{sec:basicmodel}, we describe the basic model used throughout the whole paper. 
Section \ref{sec:algorithm} deals with Bregman iteration and the operator splitting operator used in the optimization process. Section \ref{sec:imp} 
provides the whole algorithm and implementation details. Section \ref{sec:results} presents many numerical results obtained by the proposed method on real 
data. Concluding remarks are provided in section \ref{sec:conclusion}.

\section{Our Basic Model}\label{sec:basicmodel}

\begin{figure}\label{fig:deformmodel}
\centering\includegraphics[scale=0.7]{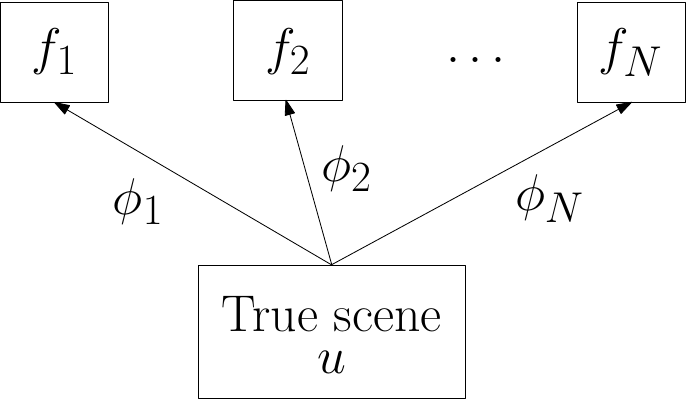}
\caption{The model of deformation used in this paper.}
\end{figure}

We denote the observed image sequence as $\{f_i\}_{i=1,\ldots,N}$ and the true image that needs to be reconstructed as $u$. We assume 
\begin{equation}\label{basic}
f_i(x)=u(\phi_i(x))+\text{noise},\quad \forall i
\end{equation}
where $\phi_i$ corresponds to the geometric deformation on the $i$-th frame (note that the $\phi_i$ are the deformations between the true image and 
the observed frame $i$ and not the continuous movement flow from frame to frame, see Fig.~\ref{fig:deformmodel}). 

If we fix $\phi_i$, $u(\cdot)\rightarrow u(\phi_i(\cdot))$ can be treated as a linear operator on $u$, so we can write the right hand side of \eqref{basic} 
as 
\begin{equation}
 u(\phi_i(\cdot))= (\Phi_i u)(\cdot)
\end{equation}
where $\Phi_i$ is the linear operator corresponding deformation $\phi_i$, then \eqref{basic} becomes 
\begin{equation}
f_i =  \Phi_iu+\text{noise}, \quad\forall i
\end{equation}
which gives the fidelity equations in the model.

On the other hand, it is reasonable to assume that our image has certain regularization features. For 
natural images the total variation  has been proved to be satisfactory for avoiding noise while preserving sharp edges \cite{Rudin:1992p2320}. 
There are many other modern models such as nonlocal TV that has been developed and investigated thoroughly 
\cite{Buades:2005p599,Gilboa:2008p2647}. If we denote the regularization term of the image as $J(u)$, then we can formulate the problem as 
\begin{equation}\label{BasicOptimization}
\min_{u,\phi_i} J(u)\quad\text{s.t.}\quad f_i=\Phi_i u+\text{noise},\quad \forall i
\end{equation}

We note that rather than investigating the time-dependent fluid dynamics model behind the warping effect, which is often extremely complicated, we simply 
treat the frames as arbitrary samples of the image after random morphing without using any sequential information of the video frames. This greatly 
simplifies the model and makes the latter available even if the sequential relevance is not strong enough in practical data. 

The regularization term of $u$, $J(u)$, has many different choices. Most notably, total variation based optimization  \cite{Rudin:1992p2320} has been 
successful for edge preserving regularization. However, the use of pure total variation models for realistic images have been shown to produce artificial 
patches; this is due to the choice of the bounded-variation space and the corresponding total variation norm. More recently, the nonlocal means 
regularization model \cite{Buades:2005p599,Gilboa:2008p2647} has been introduced which modifies the intensity of a pixel by considering the nearby pixel 
values with similar patterns. Its basic assumption is that a natural image contains repeating structures instead of repeating pixels. This method 
has proved to be successful to remove artifacts while keeping the regular pattern and texture contained in the image and has been extended to include 
variational methods using functionals with nonlocal regularization. Further it has proved to be superior to many other image regularization methods as it considers the 
large-scale structure of the image beside the local differences between pixels, which makes it capable of preserving important detailed features in an image 
while removing artifacts effectively. For these reasons, in this paper we utilize the nonlocal regularization. Thereafter, for the convenience of the reader, 
we recall the expression of $J(u)$ in the nonlocal regularization case. A detailed introduction of this regularization method and its numerical 
implementation can be found in \cite{Gilboa:2008p2647}. The Nonlocal TV (NLTV) is defined by
\begin{equation}
J(u)=J_{NLTV}(u)=\int_{\Omega}\sqrt{\int_{\Omega}\left(u(y)-u(x)\right)^2w(x,y)dy}dx
\end{equation}
where the weight $w(x,y)$ corresponds to the similarity between patches centered on pixels at $x$ and $y$. As the similarity between the patches increases, so 
does their impact on the regularized image.

There are many problems which are of a similar form to \eqref{BasicOptimization} that have been studied in recent references. For example, in 
\cite{Chan:2000p5784,Chan:2005p5787, He:2005p612}, the blind deconvolution problem is modeled as
\begin{equation}\label{blinddeconvolution}
\min_{u,k} J(u)+H(k)\quad\text{s.t.}\quad f=k\star u+\text{noise},\quad \forall i
\end{equation}
where $H(k)$ is another regularization term for the unknown convolution kernel $k$. This kind of model can be solved by the alternative optimization method, 
i.e. optimizing over different variables alternatively. We want to remark that \eqref{blinddeconvolution} is a convex problem for $u$ and $k$ respectively, 
but not convex for the joint variables, and model \eqref{BasicOptimization} has the same feature. The nature of non-convexity makes it hard to analyze the 
convergence behavior for the optimization procedure, but the alternative optimization method at least guarantees that the functional is always decreasing 
over the iterations.

In our model \eqref{BasicOptimization}, if we have a good guess on $u$, then the optimal $\phi_i$ can be estimated by \eqref{basic} via certain optical 
flow algorithms (e.g. the methods developed in \cite{Black:1996p5277,Bouguet:2000,Sun:2008p5795}). On the other hand, for fixed $\{\phi_i\}$ the model 
\eqref{BasicOptimization} can be solved as a constrained problem, which is discussed in the next section.

\section{Algorithm}\label{sec:algorithm}
\subsection{Bregman Iteration}\label{subsec:bregman}
The Bregman iterative method was originally introduced to the image processing community by \cite{Osher:2005p632}. It solves the following constrained optimization 
problem
\begin{equation}\label{Constrained}
\min_u J(u)\quad\text{s.t.}\quad f=Au+\text{noise}
\end{equation}
by solving the following series of problems
\begin{equation}\label{Bregman}
\begin{cases}
u^k=\arg\min_u J(u)+\frac{\lambda}{2}\|Au-f^k\|^2\\
f^{k+1}=f^k+f-Au^k
\end{cases}
\end{equation}
with $f^0=f$ and $A$ denotes a linear operator (the deformations in our case). It has been shown that this iteration converges to the solution of \eqref{Constrained}. 
The Bregman iterative method \eqref{Bregman} is actually equivalent to alternately descending the primal variable and ascending the dual variable of the 
Lagrangian of \eqref{Constrained}, as pointed out by many authors, e.g.  \cite{Zhang:2010p624, Zhang:2010p6848}.

\subsection{Operator Splitting Method}\label{subsec:operatorsplitting}
The first step in \eqref{Bregman} is an unconstrained problem. This formulation has appeared in many practical imaging or signal processing problems, 
see \cite{He:2005p612,Mao:2010p5528,Zhang:2010p624} for examples. It can be solved by the forward-backward operating splitting method, which  was first 
proposed by Lions and Mercier \cite{Lions:1979p606} and Passty \cite{Pas} and generalized by Combettes and Wajs \cite{Combettes:2005p617}. The scheme can 
be described as follows: to solve the unconstrained problem 
\begin{equation}\label{Unconstrained}
\min_u J(u)+\frac{\lambda}{2}\|Au-f\|^2
\end{equation}
we want to find $u$ such that $0\in\partial J(u)+\lambda A^\top(Au-f)$, where $\partial J(u)$ denotes the subdifferential of $J(u)$. This leads to the 
following fixed point algorithm:
\begin{equation}\label{operatorspliting}
\begin{cases}
v\leftarrow u-\delta A^\top(Au-f)\\
u \leftarrow \arg\min_u J(u)+\frac{\lambda}{2\delta}\|u-v\|^2
\end{cases}
\end{equation}
This method is efficient in practice. The first line, the forward step, is the gradient descent of $\|Au-f\|^2$ with step $\delta$. The second line 
\begin{equation}\label{backward}
u \leftarrow \arg\min_u J(u)+\frac{\lambda}{2\delta}\|u-v\|^2
\end{equation}
is called the backward step and can be solved efficiently for various forms of $J(u)$. For example, if we choose the total variation as the regularization term
\begin{equation}
J(u)=\int |\nabla u(x)|dx,
\end{equation}
then \eqref{backward} is the standard ROF model \cite{Rudin:1992p2320} and can be solved very efficiently via graph-cut methods \cite{Goldfarb:2009p6749}. In our 
method $J(u)$ is the nonlocal total variation, and \eqref{backward} can be solved as a typical nonlocal denoising problem as discussed in \cite{Gilboa:2008p2647}. 

Applying this method to \eqref{Bregman}, we get the following Algorithm~\ref{BregmanAlgorithm} to solve \eqref{Constrained}.

\begin{algorithm}
\caption{The Bregman Iteration}\label{BregmanAlgorithm}
\begin{flushleft}
\begin{itemize}
\item[] Initialize: Start from some initial guess $u$. Let $\tilde f=f$.\\
\item[] \textbf{while} {$\|Au-f\|^2$ not small enough} \textbf{do}\\
\begin{itemize}
\item[] \textbf{while} {$\|Au-\tilde f\|^2$ not converge} \textbf{do}\\
\begin{itemize}
\item[] $v\leftarrow u-\delta A^\top(Au-\tilde f)$\\
\item[] $u \leftarrow \arg\min_u J(u)+\frac{\lambda}{2\delta}\|u-v\|^2$\\
\end{itemize}
\item[] \textbf{end while}\\
\end{itemize}
\item[] $\tilde f \leftarrow \tilde f+f-Au$.\\
\item[] \textbf{end while}
\end{itemize}
\end{flushleft}
\end{algorithm}

\subsection{Algorithm}\label{subsec:algorithm}
We now apply Algorithm \ref{BregmanAlgorithm} to our problem. In addition to setting $\|Au-f\|_2^2=\sum_i\|\Phi_i u-f_i\|_2^2$ and $J(u)$ to be the nonlocal 
regularization, and note that in our problem $\Phi_i$ are unknown and should be updated as well. 

In our implementation we will combine the updating step for $\Phi_i$ into the Bregman updating loop, following \cite{He:2005p612}. We will choose the 
optical flow method described in \cite{Black:1996p5277} to update the $\Phi_i$. The overall algorithm is resumed on Algorithm \ref{OverallAlgorithm}.

\begin{algorithm}
\caption{The Alternative Optimization Algorithm}\label{OverallAlgorithm}
\begin{flushleft}
\begin{itemize}
\item[] Initialize: Start from some initial guess $u$. Let $\tilde f_i=f_i$.\\
\item[] \textbf{while} {$\sum_i\|\Phi_iu-f_i\|^2$ not small enough} \textbf{do}\\
\begin{itemize}
\item[] Estimate $\Phi_i$ which maps $u$ onto $f_i$ from \eqref{basic} via optical flow scheme.\\
\item[] \textbf{while} {$\sum_i\|\Phi_iu-\tilde f_i\|^2$ not converge} \textbf{do}\\
\begin{itemize}
\item[] $v\leftarrow u-\delta \sum_i\Phi_i^\top(\Phi_i u-\tilde f)$\\
\item[] $u \leftarrow \arg\min_u J(u)+\frac{\lambda}{2\delta}\|u-v\|^2$\\
\end{itemize}
\item[] \textbf{end while}\\
\item[] $\tilde f_i \leftarrow \tilde f_i+f_i-\Phi_i u$.\\
\end{itemize}
\item[] \textbf{end while}\\
\end{itemize}
\end{flushleft}
\end{algorithm}

\section{Implementation}\label{sec:imp} 

\begin{figure}[!ht]
\begin{center}
\includegraphics[width=0.32\textwidth]{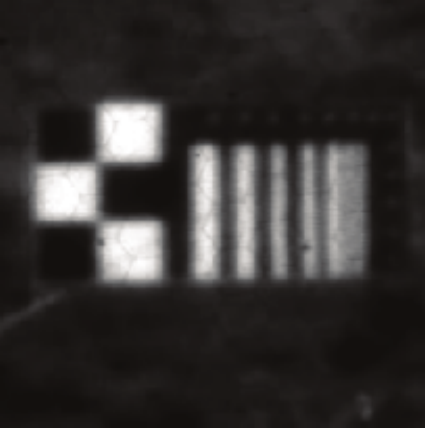}\,
\includegraphics[width=0.32\textwidth]{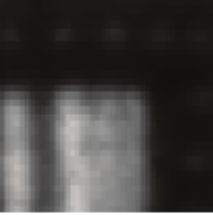}
\end{center}
\caption{The average of the frames of one example video and its magnification of the top right corner.}\label{mean}
\end{figure}

\subsection{Initial Value}\label{subsec:initial}

The initial value of $u$ is chosen as the temporal average of the frames; Fig.~\ref{mean} shows an example. We can see that the average of the frames 
is very blurry but gives a good initial guess of the rough shape of the object.

\subsection{Number of Frames}\label{subsec:num_frames}

The quantity of frames determines the quality of the reconstruction. On the other hand, the more frames used, the longer the computational time required due 
to the fact that the registration between the frames is most time consuming. In our numerical experiments we use less than 100 frames and 
generally we can always obtain satisfactory results with only 10 frames. In the synthetic example shown in Fig.~\ref{example4} we only use 20 frames.

\subsection{Computation of $\Phi_i$ and $\Phi_i^\top$}\label{subsec:warping}

As we described above, the turbulence warping $\phi_i$ is estimated using an optical flow scheme. In our implementation we use the method developed in 
\cite{Sun:2008p5795}, while other schemes can be applied as well. Given $\phi_i$, $\Phi_i u = u(\phi_i(\cdot))$ can be evaluated via interpolation. 

$\Phi_i^\top$ is the adjoint operator of $\Phi_i$, that is to say,
\begin{equation}
\langle v, \Phi_i^\top u\rangle = \langle \Phi_i v, u\rangle,\,\forall v
\end{equation}
Mathematically, this can be written as 
\begin{equation}
\int (\Phi_i^\top u)\cdot v dx = \int u\cdot v(\phi_i) dx,\,\forall v
\end{equation}
which is sometimes denoted as 
\begin{equation}
\Phi_i^\top u(x)=\phi_i\# u(x)
\end{equation}
where $\phi_i\# u(x)$ is called the push-forward action of $\phi_i$ on $u(x)$. Given $\phi_i$, $\Phi_i^\top u(x)$ can be numerically evaluated as follows:
let 
\begin{equation}
v_y(x) = \begin{cases}1, & y=x \\ 0, & y\neq x\end{cases}
\end{equation}
be the `single-pixel spike' function at pixel $y$, then the value of $\Phi_i^\top u$ at pixel $y$ can be calculated as 
\begin{equation}
(\Phi_i^\top u)(y)  = \langle v_y, \Phi_i^\top u\rangle = \langle \Phi_i v_y, u\rangle,\
\end{equation}
This computation is very efficient because $\Phi_i v_y$ is a simple function and can be directly evaluated from $\phi_i$. 

\subsection{Regularization}\label{subsec:regularization}

As discussed in section \ref{sec:algorithm} the regularization step 
\begin{equation}
u \leftarrow \arg\min_u J(u)+\frac{\lambda}{2\delta}\|u-v\|^2
\end{equation}
is a well-studied optimization problem and can be solved by many existing routines. We use the method developed in \cite{Gilboa:2008p2647} to implement 
this step. 

\subsection{Parameters}\label{subsec:parameters}

There are two parameters in our algorithm \ref{OverallAlgorithm}: $\delta$ and $\lambda$; $\delta$ is the step size for the gradient descent of the fidelity 
term. As shown in \cite{Combettes:2005p617}, the step size should be chosen such that $u\rightarrow u-\delta \sum_i\Phi_i^\top(\Phi_i u-\tilde f)$ is 
contractive to achieve the convergence. The parameter $\lambda$ is not a crucial factor and numerical results also indicate that the algorithm is not 
sensitive to the choice of $\lambda$. In practice we suggest a small initial $\lambda$ to make the image regular enough at the very beginning, then 
increase $\lambda$ gradually. This method is also used in other image reconstruction methods, e.g. \cite{Almeida:2010p6797}.

\section{Experiments}\label{sec:results}

\subsection{Tests results}
\begin{figure}[!htb]
\begin{center}
\includegraphics[width=0.45\textwidth]{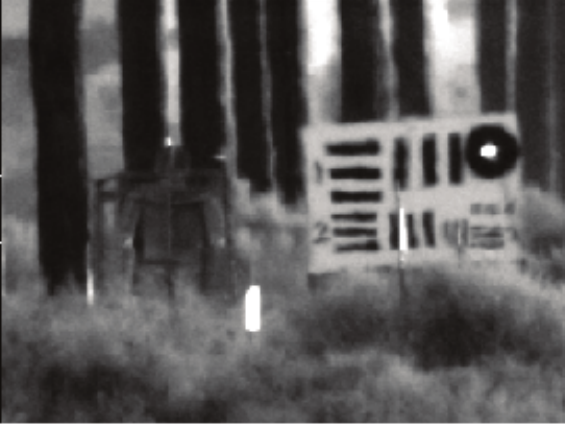}\quad
\includegraphics[width=0.335\textwidth]{frame21.pdf}\\
\includegraphics[width=0.45\textwidth]{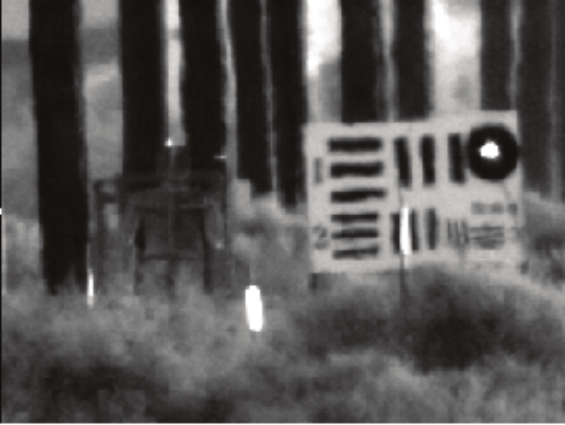}\quad
\includegraphics[width=0.335\textwidth]{frame22.pdf}\\
\includegraphics[width=0.45\textwidth]{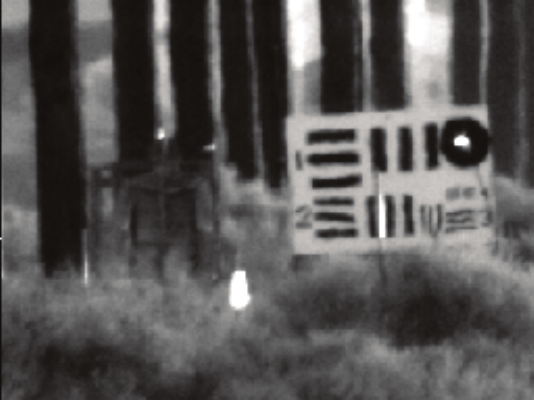}\quad
\includegraphics[width=0.335\textwidth]{frame23.pdf}\\
\includegraphics[width=0.45\textwidth]{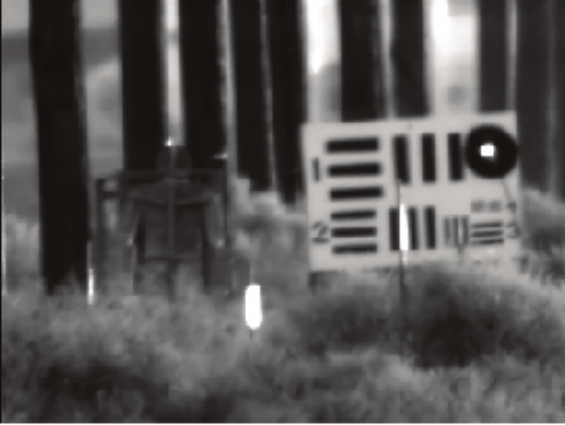}\quad
\includegraphics[width=0.335\textwidth]{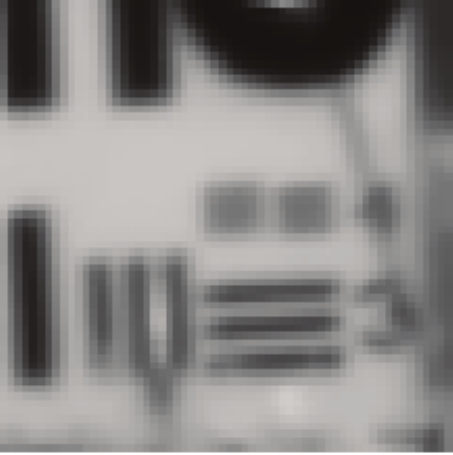}
\end{center}
\caption{The first three rows are example frames and the magnifications of the right part of the frames. The last row shows our reconstructed result.}\label{example1}
\end{figure}

\begin{figure}[!htb]
\begin{center}
\includegraphics[width=0.36\textwidth]{bigframe11.pdf}\quad
\includegraphics[width=0.36\textwidth]{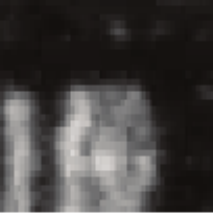}\\
\includegraphics[width=0.36\textwidth]{bigframe12.pdf}\quad
\includegraphics[width=0.36\textwidth]{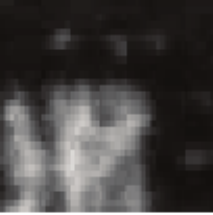}\\
\includegraphics[width=0.36\textwidth]{bigframe13.pdf}\quad
\includegraphics[width=0.36\textwidth]{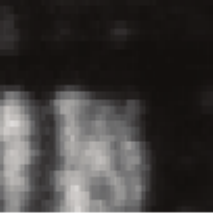}\\
\includegraphics[width=0.36\textwidth]{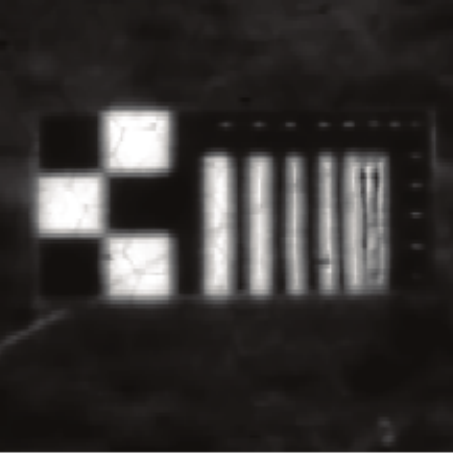}\quad
\includegraphics[width=0.36\textwidth]{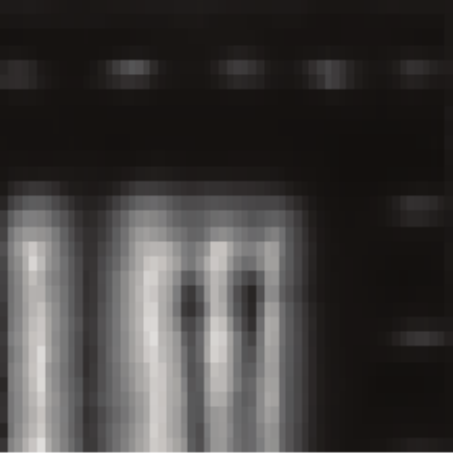}
\end{center}
\caption{The first three rows are example frames and the magnification of the top right part of the frames. The last row shows our reconstructed result.}\label{example2}
\end{figure}

\begin{figure}[!htb]
\begin{center}
\includegraphics[width=0.415\textwidth]{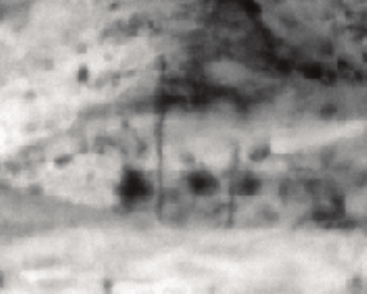}\quad
\includegraphics[width=0.33\textwidth]{frame31.pdf}\\
\includegraphics[width=0.415\textwidth]{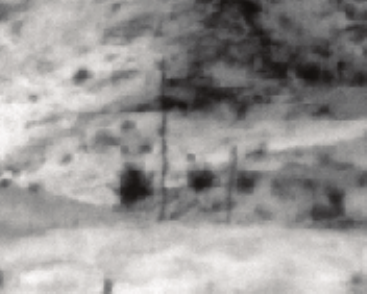}\quad
\includegraphics[width=0.33\textwidth]{frame32.pdf}\\
\includegraphics[width=0.415\textwidth]{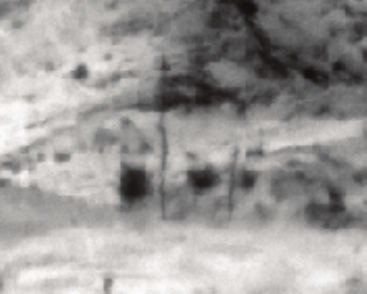}\quad
\includegraphics[width=0.33\textwidth]{frame33.pdf}\\
\includegraphics[width=0.415\textwidth]{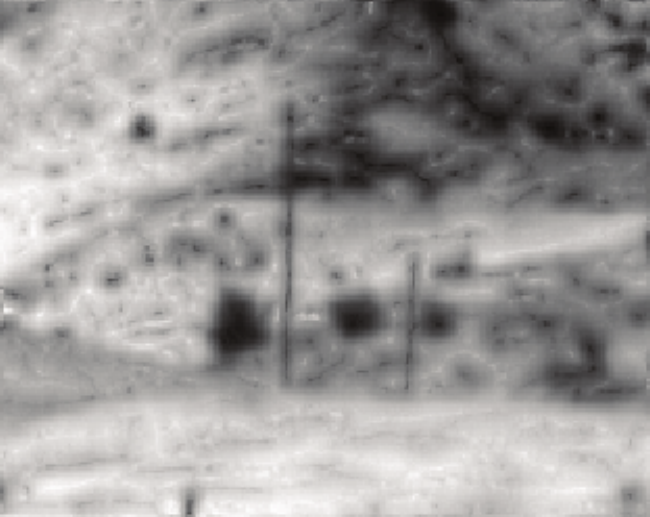}\quad
\includegraphics[width=0.33\textwidth]{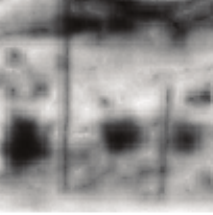}
\end{center}
\caption{The first three rows are example frames and the magnification of the center part of the frames. The last row shows our reconstructed result.}\label{example3}
\end{figure}

\begin{figure}[!htb]
\begin{center}
\includegraphics[width=1.5in]{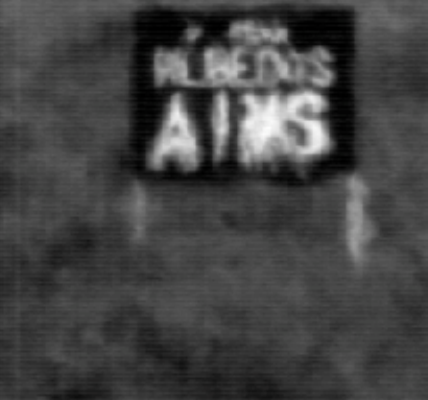}\,
\includegraphics[width=1.5in]{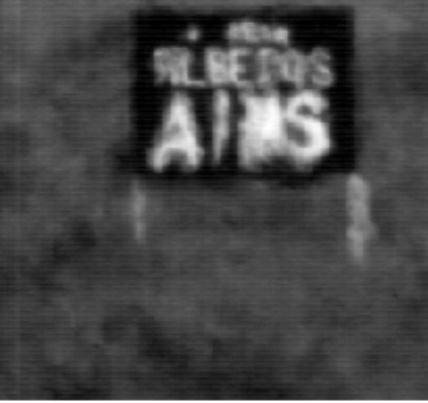}\quad
\includegraphics[width=1.5in]{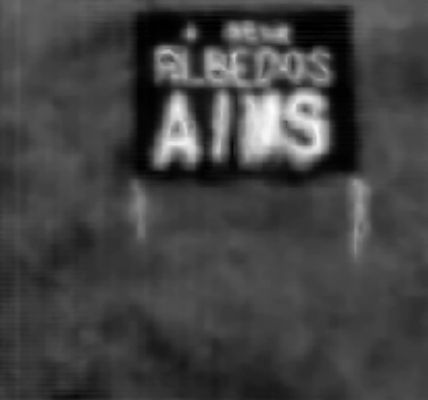}
\end{center}
\caption{The first two panels are example frames. The last panel is our reconstructed result. Only 10 frames are used in this example.}\label{example5}
\end{figure}

\begin{figure}[!htb]
\begin{center}
\includegraphics[width=1.5in]{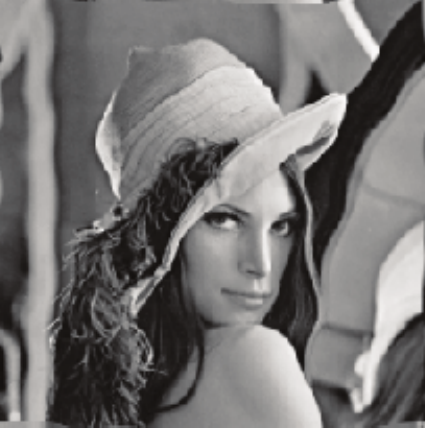}\,
\includegraphics[width=1.5in]{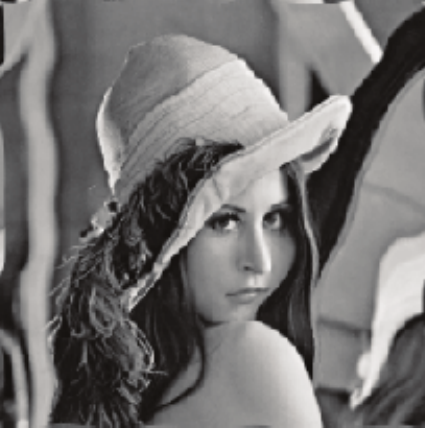}\quad
\includegraphics[width=1.5in]{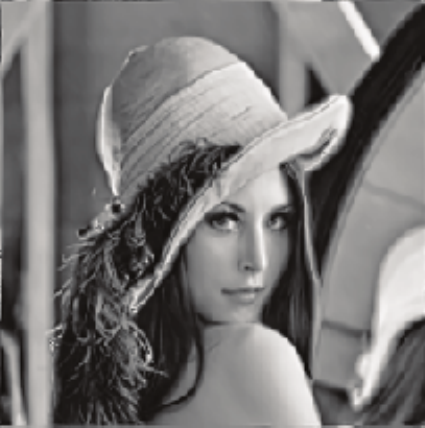}
\end{center}
\caption{The first two panels are the distorted Lena. The last figure is our reconstructed result. Only 20 frames are used in this example.}\label{example4}
\end{figure}

In Figs.~\ref{example1},~\ref{example2},~\ref{example3} and \ref{example5} some frames of test sequences from real videos are shown, as well as the magnified 
details. We ensure that all used frames are of the same quality. We check that there is no ``better'' frame in the input sequence by computing the $L^2$ 
norm between each input frames and the restored image. Our reconstruction results are shown in the last row for 
Figs.~\ref{example1},~\ref{example2},~\ref{example3} and the last column for Fig.~\ref{example5}.  Only 5 iterations are implemented in our algorithm to get 
this satisfactory result (especially in Fig.~\ref{example5} where the letters on the board are much more readable in the processed image than in the original 
frames).  Figure.~\ref{example4} shows a synthetic example, where the wave filter with unknown parameter is imposed on the static image. Only 20 frames are 
used in this example.

\subsection{Influence of the optical flow scheme choice}
\begin{figure}[!htb]
\begin{center}
\includegraphics[width=1.5in]{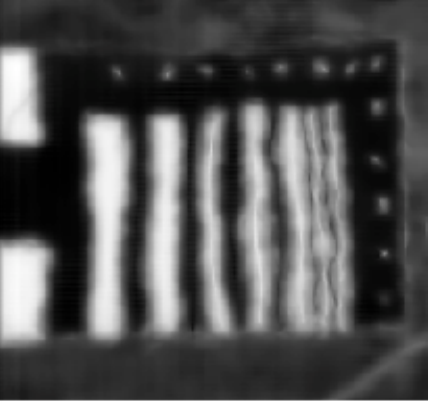}\,
\includegraphics[width=1.5in]{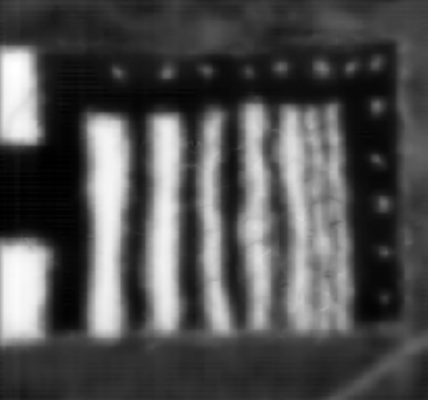}\\
\includegraphics[width=1.5in]{WSMR804_BA.pdf}\,
\includegraphics[width=1.5in]{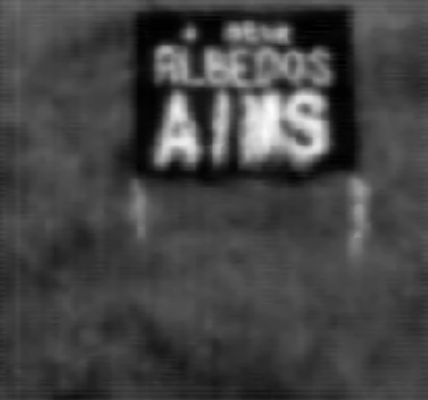}\,
\end{center}
\caption{Results obtained by using different optical flow schemes: the Black-Anandan scheme on the left column and the Lukas-Kanade scheme on the right column.}\label{example6}
\end{figure}

As previously mentioned, the optical flow scheme used was one developed in \cite{Sun:2008p5795}. This algorithm gives good optical flow estimation but is 
time consuming. In Fig.~\ref{example6}, we present results obtained by the above-mentioned algorithm and the classical Lucas-Kanade 
\cite{Bouguet:2000} optical flow algorithm which is faster. The various experiments show that the results are very similar and that our 
restoration algorithm is not sensitive to the choice of the optical flow scheme.

\subsection{The case of a nonstatic scene}
\begin{figure}[!htb]
\begin{center}
\includegraphics[width=1.5in]{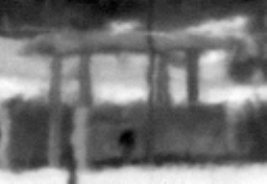}\,
\includegraphics[width=1.5in]{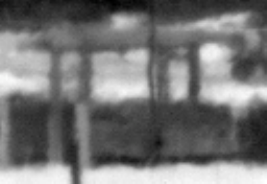}\,
\includegraphics[width=1.5in]{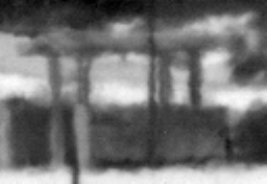}\\
\includegraphics[width=1.5in]{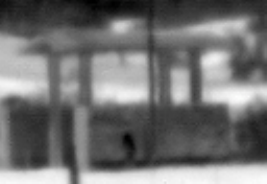}\,
\includegraphics[width=1.5in]{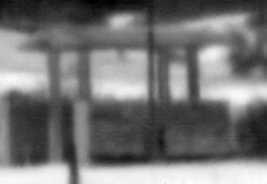}\,
\includegraphics[width=1.5in]{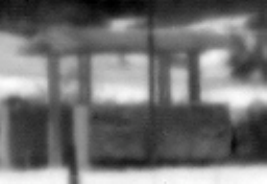}\,
\end{center}
\caption{Results obtained on a sequence with a moving pedestrian (original frames on top and restored frames on bottom).}\label{example7}
\end{figure}

In Fig.~\ref{example7}, we show some results when the proposed algorithm is applied on a sequence in which a pedestrian is moving. A restored frame $N$ is 
processed with only ten previous frames $\{N-10,\ldots,N\}$. The number of frames used by the algorithm clearly depend on the velocity difference between 
the movement of the pedestrian and movements due to turbulence. The results show an improved sequence where the turbulence deformations are mostly 
compensated without altering the pedestrian movement which will be easier to detect.

\subsection{Comparison with existing methods}
\begin{figure}[!htb]
\begin{center}
\includegraphics[width=1.5in]{WSMR804_BA.pdf}\,
\includegraphics[width=1.5in]{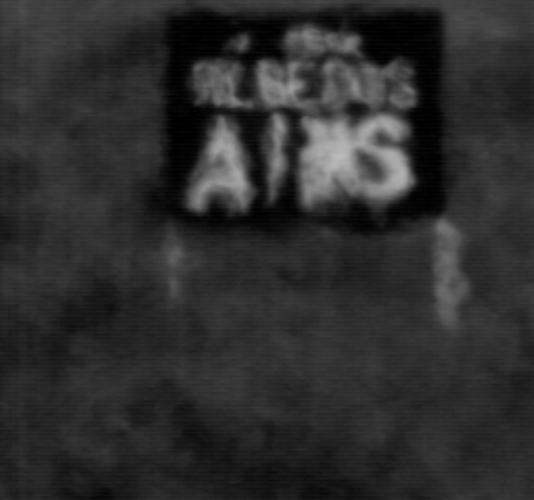}\,
\includegraphics[width=1.5in]{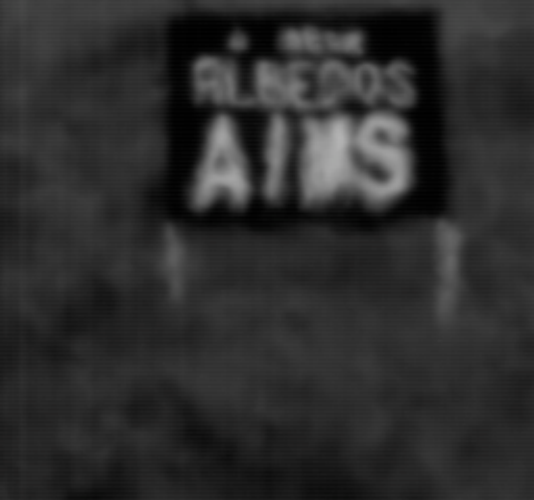}\\
\includegraphics[width=1.5in]{WSMR856_N100BA.pdf}\,
\includegraphics[width=1.5in]{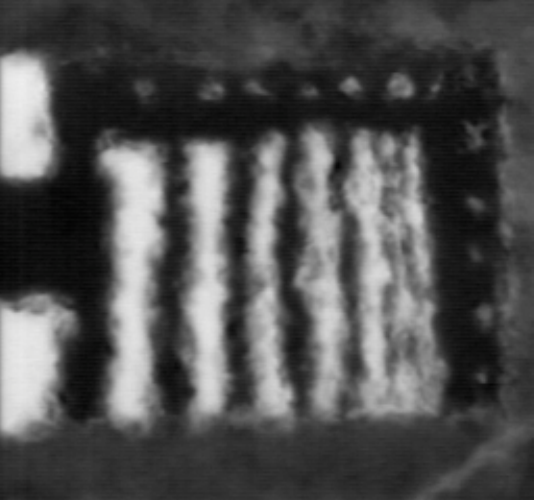}\,
\includegraphics[width=1.5in]{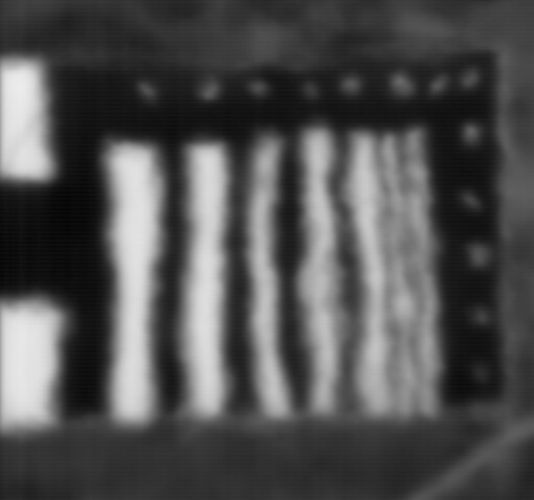}\,
\end{center}
\caption{Results obtained with different algorithms: our method on left, the PCA based approach in the middle and the Lucky-Region Fusion on the right.}\label{example8}
\end{figure}

Figure.~\ref{example8} shows a comparison between results obtained using our method and two other state of the art methods: the algorithm based on PCA 
\cite{Li2007} and the algorithm using the Lucky-Region Fusion \cite{Aubailly2009}. In these tests, we deal with short exposure sequences; it is not 
surprising that the PCA method failed as in this case the assumption of a gaussian kernel for the blur is not correct. The Lucky-Region Fusion approach 
gives some good results but our method shows a better geometry reconstruction (see, for example the high spatial frequencies in the second row, the 
separation between each bar is clearer in our results). The other advantage of our algorithm is that it requires only a few frames to get good results, compared 
to 100 frames which were used in the Lucky-Region Fusion.

\section{Conclusion}\label{sec:conclusion}
We propose a novel approach to restore an image from a video sequence contaminated by geometric deformations like the ones observed in atmospheric 
turbulence. Our method is based on a variational model solved by Bregman Iterations and the operator splitting method. Both state of the art and classical 
optical flow methods were utilized to estimate the warping effect. No restriction on the frame rate or sequence order is needed in our method. The algorithm 
is simple, concise and computationally efficient.

In order to deal with the complete turbulence problem and following the model of Frakes \cite{Frakes:2001p7017}, we try some deblurring at the end of the 
process. We test both nonblind deconvolution (by assuming a gaussian blur kernel, where its size is chosen experimentally) based on total variation 
regularization \cite{Goldstein2009} or framelet decomposition \cite{Cai,Cai2009b}, and Bayesian blind deconvolution (the \textit{deconv} function available 
in Matlab).

\begin{figure}[!htb]
\begin{center}
\includegraphics[width=1.5in]{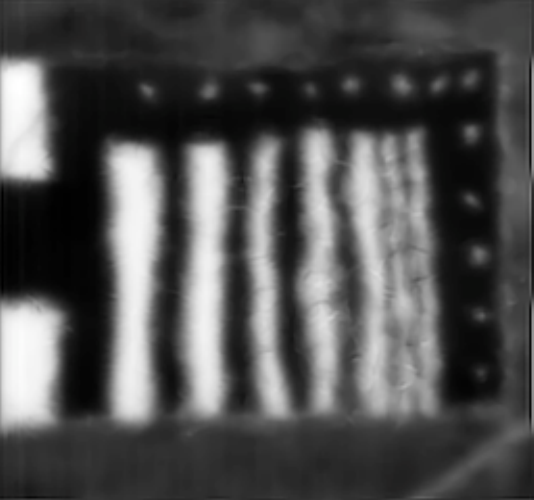}\,
\includegraphics[width=1.5in]{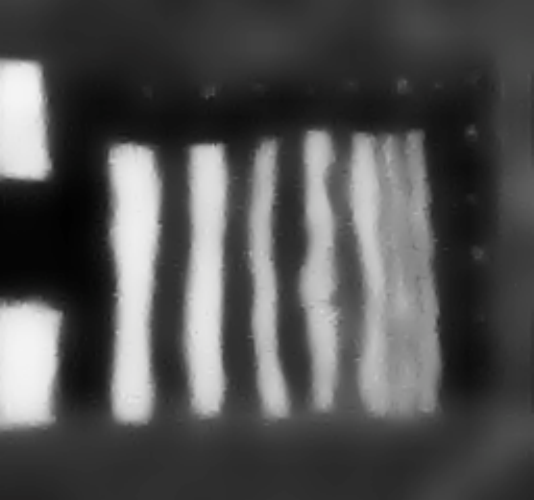}\,
\includegraphics[width=1.5in]{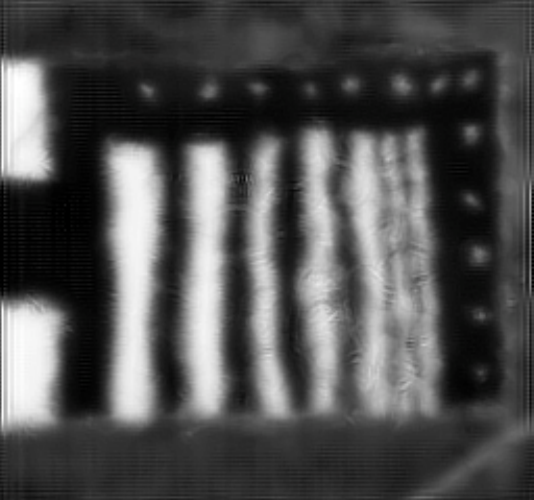}\\
\includegraphics[width=1.5in]{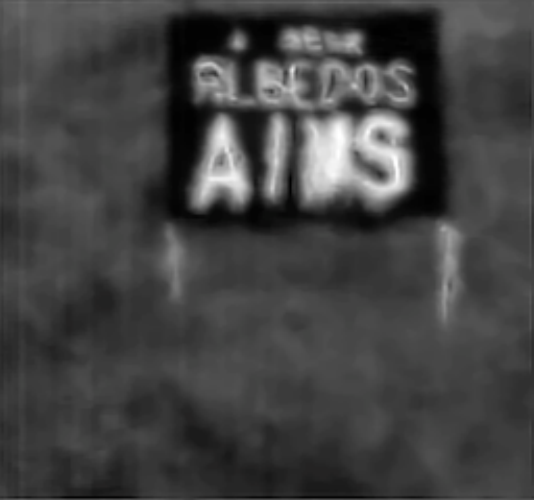}\,
\includegraphics[width=1.5in]{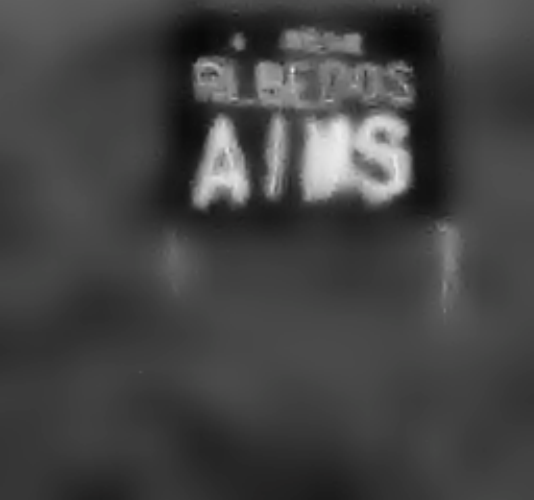}\,
\includegraphics[width=1.5in]{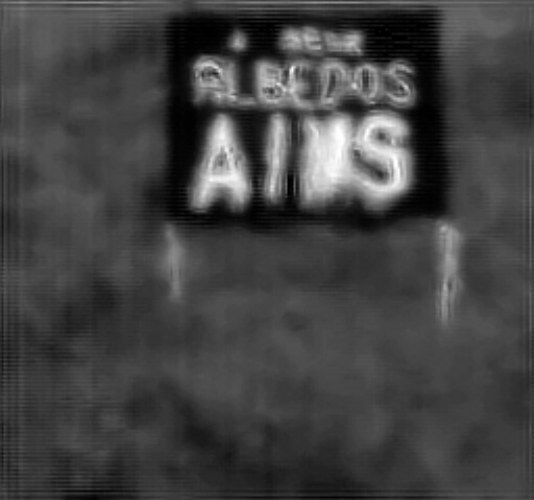}\,
\end{center}
\caption{Outputs of three different deblurring algorithms: Nonblind TV-deblurring on left, a Nonblind Framelet based approach in the middle and the Matlab Bayesian Blind deconvolution on right.}\label{example9}
\end{figure}

The results, shown in Fig.~\ref{example9}, obtained by adding these deblurring effects don't show much improvement. For the two nonblind techniques 
the differences between the input and output are rather small. Further, different experiments show that the Gaussian assumption for 
the kernel is not correct for this kind of blur. The Bayesian blind algorithm gives images with sharper edges, but as previously shown, the improvements are not 
really significant. We are currently investigating the best way to deblur. The main questions are: which kind of deblurring method to use? When is the best 
moment to apply deblurring (at the end of the process or during the Bregman iteration)?

Finally, we aim to explore the broader capabilities of this method in different applications like underwater imaging.
 
\section*{Acknowledgment}
The authors want to thank the members of the NATO SET156 (ex-SET072) Task Group for the opportunity of using the data collected by the group during 
the 2005 New Mexico's field trials, and the Night Vision and Electronic Sensors Directorate (NVESD) for the data provided during the 2010 summer 
workshop. The authors thank Prof. Jean-Michel Morel at ENS de Cachan and Prof. Stanley Osher and Dr. Yifei Lou at UCLA for helpful discussions. 
This work is supported by the following grants: NSF DMS-0914856, ONR N00014-08-1-119, ONR N00014-09-1-360.

\medskip

Received July 2011.

\medskip
{\it E-mail address: }maoxa004@ima.umn.edu\\
\indent{\it E-mail address: }jegilles@math.ucla.edu\\

\end{document}